\documentclass{article}
\usepackage{PRIMEarxiv}

\usepackage[utf8]{inputenc} 
\usepackage[T1]{fontenc}    
\usepackage{hyperref}       
\usepackage{url}            
\usepackage{booktabs}       
\usepackage{amsfonts}       
\usepackage{nicefrac}       
\usepackage{microtype}      
\usepackage{lipsum}
\usepackage{fancyhdr}       
\usepackage{graphicx}       
\graphicspath{{media/}}     
\usepackage{float}

\usepackage{xspace}
\usepackage{amssymb}
\usepackage{booktabs}
\usepackage{multirow}
\usepackage{arydshln}
\usepackage{amssymb}
\usepackage{hhline}
\usepackage{tcolorbox}
\usepackage{verbatim}
\usepackage{listings}
\usepackage{fancyvrb}
\usepackage{xcolor}

\newcommand{\coder}{\textsc{CodeR}\xspace}
\newcommand{\swebench}{SWE-bench\xspace}
\newcommand{\swebenchlite}{SWE-bench lite\xspace}
\newcommand{\sweagent}{SWE-agent\xspace}
\newcommand{\autocoderover}{AutoCodeRover\xspace}

\definecolor{darkgreen}{rgb}{0.0, 0.5, 0.0}

\title{\coder: Issue Resolving with Multi-Agent and\\
Task Graphs}

\author{
Dong Chen\textsuperscript{1}\thanks{Equal contribution}
~~Shaoxin Lin\textsuperscript{1\textasteriskcentered{}}~~Muhan Zeng\textsuperscript{1\textasteriskcentered{}}~~Daoguang Zan\textsuperscript{2\textasteriskcentered{}} \\ 
\textbf{Jian-Gang Wang\textsuperscript{1}}~~\textbf{Anton Cheshkov\textsuperscript{1}}~~\textbf{Jun Sun\textsuperscript{3}}~~\textbf{Hao Yu\textsuperscript{4}}~~\textbf{Guoliang Dong\textsuperscript{3}}~~\textbf{Artem Aliev\textsuperscript{1}}
\\
\textbf{Jie Wang\textsuperscript{1}}~~\textbf{Xiao Cheng\textsuperscript{1}}~~\textbf{Guangtai Liang\textsuperscript{1}}~~\textbf{Yuchi Ma\textsuperscript{1}}~~\textbf{Pan Bian\textsuperscript{1}}~~\textbf{Tao Xie\textsuperscript{4}}~~\textbf{Qianxiang Wang\textsuperscript{1}}
\\
\textsuperscript{$^1$}Huawei Co., Ltd.
\textsuperscript{$^2$}Chinese Academy of Science 
\textsuperscript{$^3$}Singapore Management University
\textsuperscript{$^4$}Peking University
}

\begin{document}
\maketitle

\begin{abstract}
GitHub issue resolving recently has attracted significant attention from academia and industry.
\swebench~\cite{swe-bench} is proposed to measure the performance in resolving issues.
In this work, we propose \coder, which adopts a multi-agent framework and pre-defined task graphs to \textbf{R}epair \& \textbf{R}esolve reported bugs and add new features within code \textbf{R}epository.
On \swebenchlite, \coder is able to solve $28.33\%$ of issues, when submitting only once for each issue.
We examine the performance impact of each design of \coder and offer insights to advance this research direction\footnote{\url{https://github.com/NL2Code/CodeR}}.

\end{abstract}

\section{Introduction}

The rapidly growing capability of Large Language Models (LLMs) is dramatically reshaping many industries~\cite{nl2code,zhang2023survey,zheng2023survey}. 
The most recent release of GPT-4o~\cite{gpt4o} demonstrates a significant leap in multi-modal capabilities and artificial intelligence (AI)-human interaction, whilst maintaining the same level of text generation, reasoning, and code intelligence as GPT-4-Turbo~\cite{gpt4}.
LLMs can interact with humans and the world as humans do, it is considered a starting point for LLMs to take over tasks from humans or collaborate naturally with humans. 

Issue resolving is one of the software engineering tasks experimented with LLMs that is particularly relevant in practice.
SWE-bench~\cite{swe-bench} collects $2,\!294$ real-world issues from $12$ popular Python libraries. 
The LLMs are tasked to resolve the issues based on the given issue description and the whole repository.
This task is extremely challenging due to the need for deep reasoning about a huge amount of code and incomplete information for the task description. 
SWE-bench-lite~\cite{swe-bench} removes the issue with low-quality descriptions to make the task more addressable, and yet it remains highly non-trivial.

Since SWE-bench was released, multiple approaches have been proposed.
SWE-Llama~\cite{swe-bench} adopt a pipeline with Retrieval-Augmented Generation (RAG) to generate the patch directly. 
Later, \autocoderover~\cite{autocoderover} added code contextual retrieval with keywords in the issue description into the pipeline. 
It iteratively collects code context by the keywords in the issues until LLMs have collected enough information to generate a correct patch.
Instead of explicitly patch generation, \sweagent~\cite{yang2024sweagent} performs iterative edits in the repository.
It then uses the ``\texttt{git diff}'' command to generate patches which avoids patch format errors. 

In the literature on applying LLMs for solving software engineering tasks, multiple agent-based approaches have shown their competitiveness.
For instance, MetaGPT~\cite{metagpt} uses the multi-agent approach to automate the software development process from scratch. \autocoderover~\cite{autocoderover} and \sweagent~\cite{yang2024sweagent} use the single-agent approach to address automatic GitHub issue resolving.

To the best of our knowledge, in issue resolving scenarios, the agent-based approaches primarily focus on a single agent.
Moreover, previous works perform task decomposition on-the-go, with each subsequent step being determined by the preceding one.
Multi-agent possesses the advantage of better decoupling each role and leveraging contextual information.
However, implementing a multi-agent framework in issue resolving presents challenges such as: 
(1) Free communications between agents may lead to a non-progressing loop without termination~\cite{wen2019modelling}. 
(2) Information passed from one agent to another may incur information loss~\cite{su2019improving}.
(3) Complex plans are hard to follow when multiple agents are involved.
We remark that these problems are not unlike those when human developers collaborate.
In this work, we develop a multi-agent design called \coder that effectively addresses the above mentioned problems.

\coder adopts a multi-agent framework and a task graph data structure for issue resolving tasks. Our design is based on the following intuitions:
\begin{itemize}
\itemindent=-15pt
\item \textbf{Less candidate actions, easier decision.} 
We introduce a set of diverse actions for different purposes. 
The number of actions is much larger compared with the single-agent framework such as SWE-agent. 
To address the problem of the large number of actions, we reduce the complexity of making decisions for the next action by limiting each agent's focus to a subtask and a subset of associated actions.
\item \textbf{Look before you leap.} We believe that planning at the beginning of the pipeline is better than deciding the next steps on-the-go. Moreover, a good plan should consist of small and manageable tasks that LLMs were trained to solve.
\item \textbf{Bypassing instruction-following and memorization.} The conventional plan generated by LLM is in the form of plain text. 
It is usually placed in the prompt to guide the subsequent steps in a LLM-centered system. 
It requires the LLM to have a strong instruction-following ability and to have a ``good'' memory to execute the plan precisely and iteratively. 
For complex tasks, like issue resolving with complex tools, task plans in pure-text prompts will be hard to follow.
Therefore, we introduce a new data structure namely \emph{task graph} that can ensure that all pre-designed plans are accurately followed and executed.
\end{itemize}

Our contributions are as follows:
\begin{enumerate}
\itemindent=-15pt        
\item We propose \coder, a multi-agent framework with task graphs for issue resolving. 
Inspired by the issue resolving process by humans in the real world, we design the roles and the actions.
For plans, we design a graph data structure that can be parsed and strictly executed. 
It can ensure the exact execution of the plan and at the same time provide an easy-to-plug interface for plan injection from humans.
\item We leverage LLM-generated code for reproducing the issue and the tests in the repository (excluding the verification tests) to get code coverage information. 
Coverage information improves contextual retrieval based on the keywords in the issue text and does fault localization together with BM25. 
\item We renew the state-of-the-art of \swebenchlite to $28.33\%$ ($85$/$300$) with only one submission per issue.
\end{enumerate}

\section{Framework}

\begin{figure}[h!]
    \centering
    \includegraphics[width=1\textwidth]{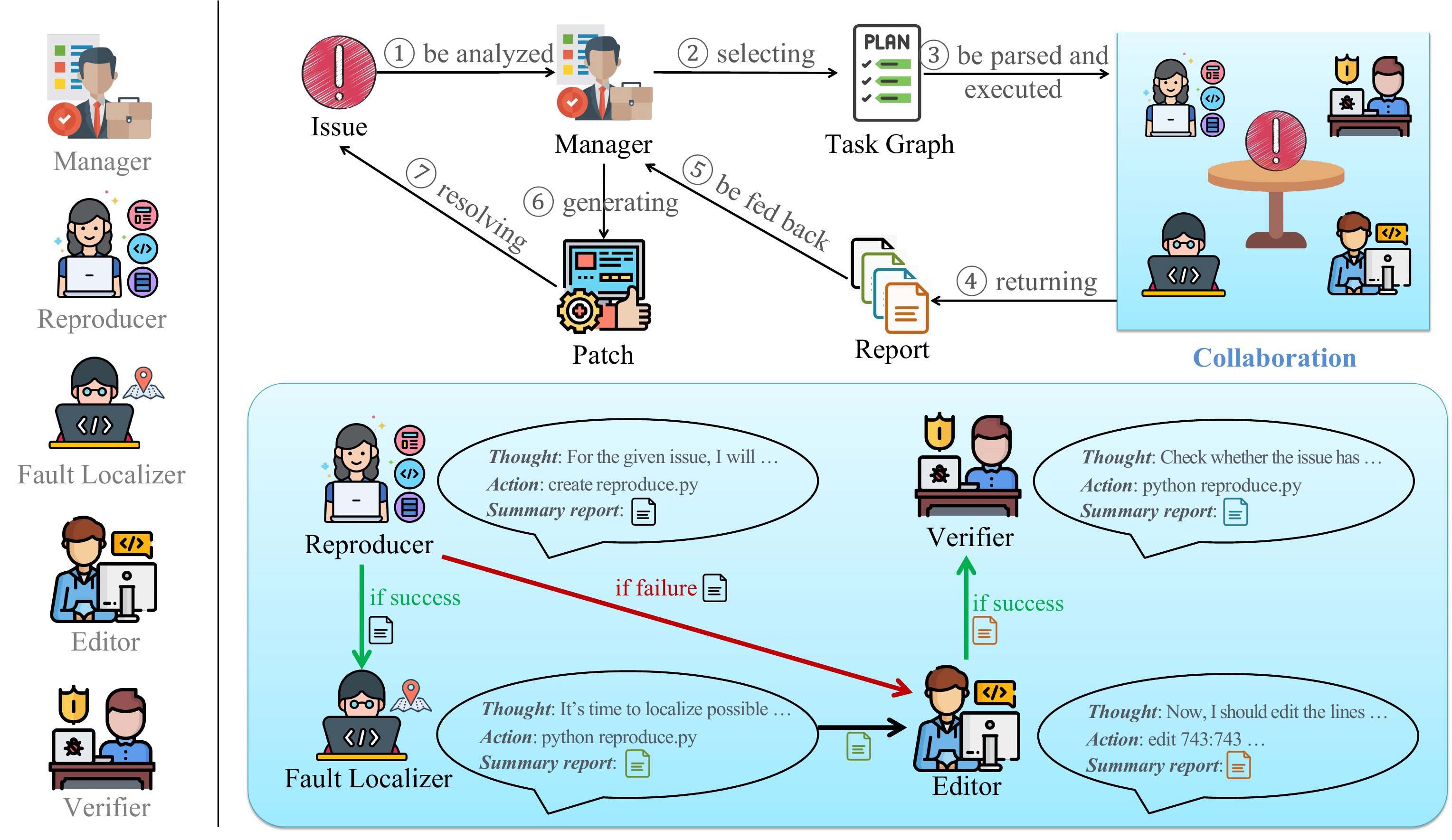}
    \caption{Multi-Agent framework of \coder with task graphs.}
    \label{fig:multi-agent}
    \vspace{-0.3cm}
\end{figure}

As Figure ~\ref{fig:multi-agent} shows, 
our design contains five agents, which can collaboratively solve GitHub issues:
\begin{itemize}
  \itemindent=-15pt
  \item Manager: The manager is an agent who interacts with the user directly and is in charge of the whole issue-resolving task. It has two responsibilities: 
  (1) selecting a plan according to the issue description. The plan specifies the agents evolved and how they should interact to finish the task.
  (2) interpreting the execution summary of a plan. If the execution summary has indicated that the issue has been solved, it will summarize the changes and submit a patch; if not, it will come up with a new plan or give up.
  \item Reproducer: The reproducer is an agent that is responsible for generating a test to reproduce the issue. 
  If the issue description contains a complete test,
  the reproducer only needs to copy the test into a new test file ``reproduce.py'', and execute and compare the output. But this is usually not the case for real-world issues, the reproducer often needs to adjust or generate test cases. We generate test cases by extracting test inputs from issues and using LLMs to generate test sequences.
  \item Fault Localizer: The fault localizer is an agent that identifies the code regions that could cause the issue. It is equipped with several fault localization tools in software engineering.
  \item Editor: The editor is the one who performs the actual code changes. It will utilize all information provided by other upstream agents and will gather contextual information with \autocoderover's search~\cite{autocoderover}. With enough information gathered, the iterative edits same as SWE-agent will be performed~\cite{yang2024sweagent}.
  \item Verifier: The verifier is an agent that will run the reproduced or integration tests\footnote{Integration tests refer to those built-in unit tests in the repository rather than official issue tests of \swebenchlite.} to check whether the modifications have resolved the issue or not.
\end{itemize}

For actions, we reuse the actions that are defined by \sweagent and \autocoderover as Table~\ref{tab:actionTable} shows. 
Besides, we also introduce new actions $0$ and $18$-$21$.
Action~$0$ selects or generates feasible plans by analyzing the current issue.
Action~$18$ retrieves the top-1 similar issue and its corresponding patch by description. 
Note that we prompt the agent to check whether the retrieved result is relevant to the current issue and analyze how its patch solves the retrieved issue.
Action~$19$ performs fault localization described in Section~\ref{sec:fl}.
Action~$20$ runs the reproducer-generated test and the integration tests. 
Same as Aider, the integration tests do not contain the tests to verify the correctness of the generated patches~\cite{aider}.
Action~$21$ summarizes all actions performed and observations by each agent for a sub-task.
Action~$22$ provides basic Linux shell commands such as ``\texttt{cd}'', ``\texttt{ls}'', ``\texttt{grep}'', and ``\texttt{cat}''.

We assign a unique set of actions to each role, similar to how different roles in the real world possess distinct skills.
For example, only the Manager has the permission to the ``\texttt{plan}'' and ``\texttt{submit}'' actions;
All roles are granted permission to use the ``\texttt{basic shell commands}'' action.

\begin{table}[t!]
    \centering
    \caption{Actions selected and designed for each agent. 
    1-10 are from SWE-agent and 11-17 are from AutoCoderRover.
    $*$~indicates that actions 11-17 are the enhancement versions of \autocoderover's original actions described in Section~\ref{sec:fl}.
    }
    \label{tab:actionTable}
    \begin{tabular}{l|ccccc} 
\toprule
\multirow{2}{*}{\textbf{Actions}} & \multicolumn{5}{c}{\textbf{Agent Roles}}                                                                                     \\ 
\cdashline{2-6}
                                  & \multicolumn{1}{l}{\textbf{Manager}} & \textbf{Reproducer} & \textbf{Fault Localizer} & \textbf{Editor} & \textbf{Verifier}  \\ 
\hline\hline
0 plan                            &               $\surd$                       &              &                          &          &                    \\
\cdashline{1-1}
1 open                            &                                      & $\surd$             &                          & $\surd$         &                    \\
2 goto                            &                                      & $\surd$             &                          & $\surd$         &                    \\
3 scroll down                     &                                      & $\surd$             &                          & $\surd$         &                    \\
4 scroll up                       &                                      & $\surd$             &                          & $\surd$         &                    \\
5 create                          &                                      & $\surd$             &                          & $\surd$         &                    \\
6 edit                            &                                       & $\surd$             &    $\surd$                       & $\surd$         &          $\surd$           \\
7 submit                          & $\surd$                              &                     &                          &                 &                    \\
8 search dir                      & $\surd$                              & $\surd$             &                          & $\surd$         &                    \\
9 search file                     & $\surd$                              & $\surd$             &                          & $\surd$         &                    \\
10 find file                      & $\surd$                              & $\surd$             &                          & $\surd$         &                    \\ 
\cdashline{1-1}
11 rover search file$^*$              & $\surd$                              & $\surd$             &                          & $\surd$         &                    \\
12 rover search class$^*$             & $\surd$                              & $\surd$             &                          & $\surd$         &                    \\
13 rover search class in file$^*$     & $\surd$                              & $\surd$             &                          & $\surd$         &                    \\
14 rover search method$^*$            & $\surd$                              & $\surd$             &                          & $\surd$         &                    \\
15 rover search method in file$^*$    & $\surd$                              & $\surd$             &                          & $\surd$         &                    \\
16 rover search code$^*$              & $\surd$                              & $\surd$             &                          & $\surd$         &                    \\
17 rover search code in file$^*$      & $\surd$                              & $\surd$             &                          & $\surd$         &                    \\ 
\cdashline{1-1}
18 related issue retrieval    &                                      &                     &              $\surd$         & $\surd$         &                    \\
19 fault localization             &                                      &                     & $\surd$                  &                 &                    \\
20 test                        &                                      &                     &                          &                 & $\surd$            \\
21 report                    &                                      & $\surd$             & $\surd$                  & $\surd$         & $\surd$                \\
\cdashline{1-1}
22  basic shell command                &    $\surd$                                  & $\surd$             & $\surd$                  & $\surd$         & $\surd$                \\
\bottomrule
\end{tabular}
\end{table}

\section{Methodology}

Repository-level tasks usually require processing a huge amount of information and taking many steps before reaching their desired solutions. 
Existing works show that dividing a repository-level task into a set of connected sub-tasks and conquering them one by one could be effective. Parsel~\cite{zelikman2023parsel} and CodeS~\cite{zan2024codes} focus on generating a large piece of code for complex algorithms and simple repositories. 
Both of them utilize inherent program structures like call graphs or file structures for task decomposition.
Issue resolving is also a repository-level task but is closer to a modification task rather than a generation task.
In addition to generating code, a repository-level modification task requires identifying the correct locations before generating the correct code. It is unfeasible to use the whole repository as input context.
This introduces additional steps and complexity which requires a more powerful framework for planning. 

\subsection{Task Graphs for Planning}

The description of GitHub issues is extremely diverse. Some issues only have one sentence in natural language (e.g. astropy\_\_astropy-7008\footnote{\url{https://github.com/astropy/astropy/pull/7008}}). 
Some may provide the test code, running results of the test code, and a possible solution (sympy\_\_sympy-14774\footnote{\url{https://github.com/sympy/sympy/pull/14774}}).
Besides descriptions, the solutions of issues are also varied.
Some could only require changing one or two lines to resolve, making the task similar to a line completion task with context (scikit-learn\_\_scikit-learn-13779\footnote{\url{https://github.com/scikit-learn/scikit-learn/pull/13779}})
while some could necessitate changing multiple files, requiring a deep understanding of the code semantics within the repository.

For simple issues with clear descriptions, their solutions are obvious and can be figured out at first glance. But for complex ones with ambiguous or inaccurate descriptions, executing tests and searching through the code base or web could be beneficial for solving them.
To cope with different approaches to solving an issue, we design a task graph that can easily add new plans. It can also be strictly followed by multi-agent systems. 

\begin{figure}[H]
    \centering
    \includegraphics[width=1.0\textwidth]{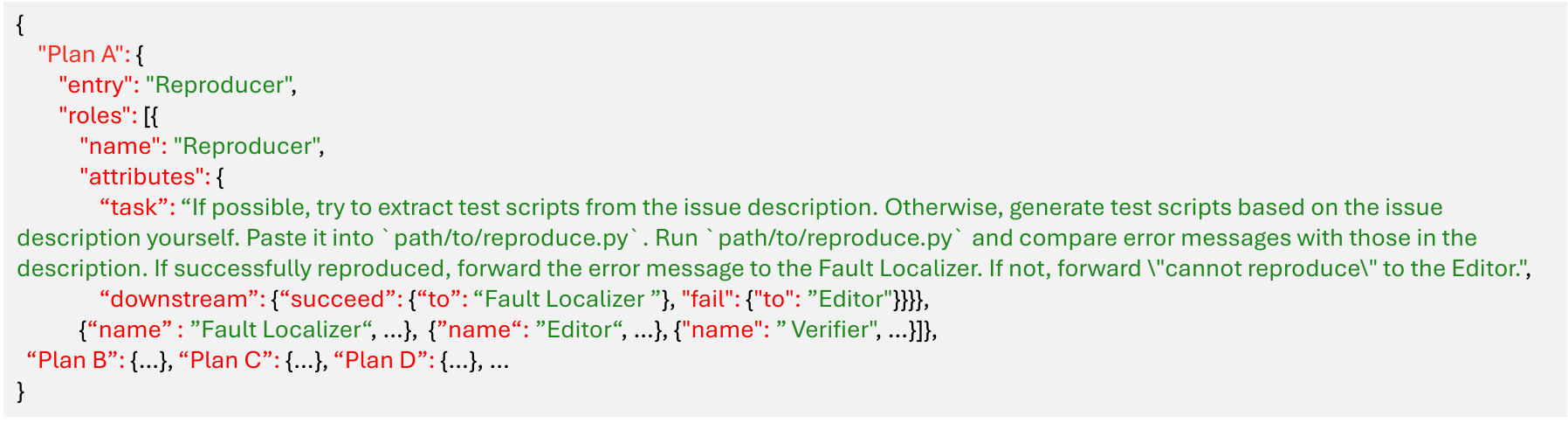}
    \caption{Task graphs in JSON format.}
    \label{fig:plans_prompt}
\end{figure}

Figure~\ref{fig:plans_prompt} shows a task graph plan in JSON format. It specifies a collection of plans in the top level with the name ``Plan ID''. For each plan, ``entry'' specifies which agent to start with. ``roles'' specifies a list of agents that are involved in this plan. Each selected agent will be given a subtask specified in ``task''. Once finished, all actions that the agent performed will be summarized and passed to its ``downstream'' according to the result of the current sub-task.
Plan A in Figure~\ref{fig:plans_prompt} involves four agents: Reproducer, Fault Localizer, Editor, and Verifier. This plan starts with Reproducer as demonstrated in Figure~\ref{fig:multi-agent}.

This design of plans decouples agent design with the task decomposition.
When designing the agents, one can only focus on the high-level goal of a sub-task without considering the details of the diverse approaches. 
The diversity of approaches can be specified and adjusted in the field of ``task'' and ``downstream''. 
In this way, the plans can be easily added, deleted, and tuned without changing a single line of code for agents.

Plans in Figure~\ref{fig:plans_prompt} will be parsed into a graph with an entry node specified by ``entry''. When starting to execute the plan, the entry node is activated and the specified agent will start to execute its sub-task using the ReAct framework~\cite{yao2022react} iteratively.
Once finished with its subtask, it will activate one of its specified ``downstream'' nodes. 
Agents in the plan may be activated multiple times if there is a cycle in the plan.
The plan finishes when the Manager is activated or exceeds our budget. 

\begin{figure}[H]
    \centering
    \includegraphics[width=\textwidth]{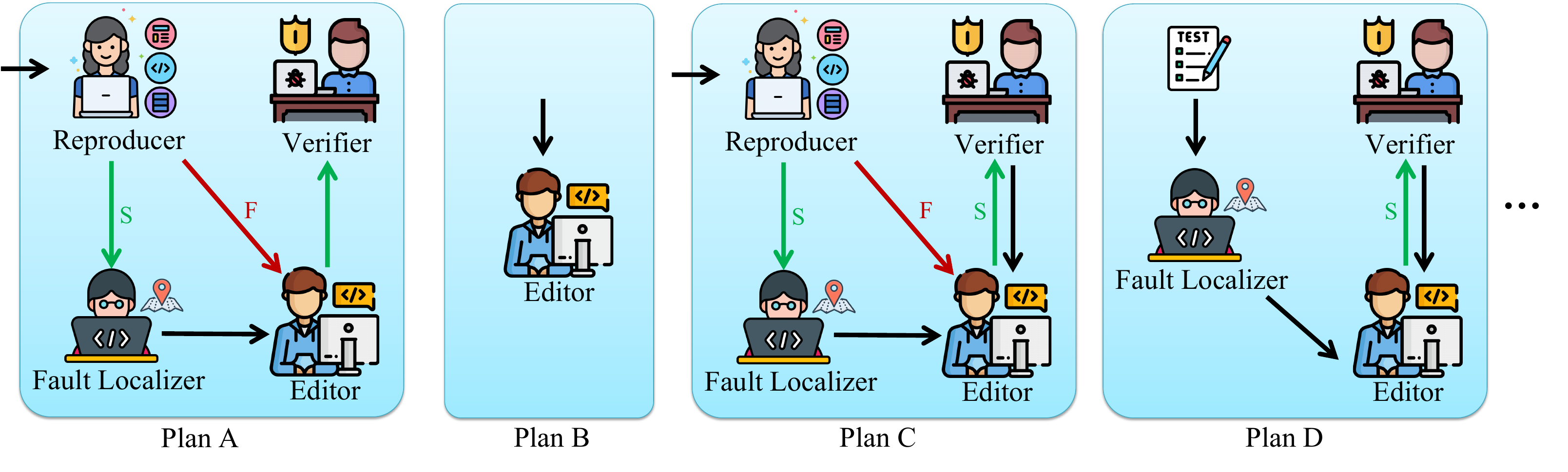}
    \caption{Plans in the form of structured graphs. They will be parsed into a graph when executed. The \textcolor{darkgreen}{\textbf{green}} and \textcolor{red}{\textbf{red}} arrows represent the reports passed to the next agent in cases of \textcolor{darkgreen}{\textbf{Success}} and \textcolor{red}{\textbf{Failure}}, respectively.
    The \textbf{black} arrows indicate the reports are passed to the next agent regardless of success or failure.}
    \label{fig:plans_details}
\end{figure}

We have designed four plans as Figure~\ref{fig:plans_details} shows. 
Plan A is shown in Figure~\ref{fig:multi-agent}, which is a standard flow to resolve an issue. It has no loop for simplicity and robustness. 
Plan B tries to resolve the issue directly for simple issues.
Plan C adds a loop that allows the feedback from testing. This circle is also used by Aider~\cite{aider} with tests that are not related to the issue (which is also called ``\texttt{integration tests}'').
Plan D takes a test-driven approach with a ground truth test for issues (such as ``\texttt{fail-to-pass}'' and ``\texttt{pass-to-pass}'' tests in \swebench).
In our experiments, we use only Plan A and B for cost savings and fast evaluation.
 
\subsection{Fault Localization Specialized for Issue Resolving}
\label{sec:fl}
We leverage fault localization techniques~\cite{fl-survey} to provide precise location information. 
A previous work~\cite{autocoderover} shows that the use of fault localization techniques leads to an increase in the efficacy of resolving GitHub issues.

We notice that the agent is allowed to run test suites but only the results are used while runtime information is not captured during the process. Test-based fault localization can provide precise location information based on runtime information and specifically, we use spectrum-based fault localization (SBFL) as the main fault localization method.

SBFL is a lightweight, test-based fault localization technique. Given a test suite that contains at least one failing test, SBFL collects statement coverage for the test suite. Suspiciousness score is then calculated based on coverage data, and all covered statements are ranked by their suspiciousness. Suspiciousness score can be calculated by different formulas such as Ochiai~\cite{ochiai} and Tarantula~\cite{tarantura}. These formulas share the same motivation that the fault location should possibly be covered by more failing tests and fewer passing tests.

One main limitation of SBFL and many other test-based fault localization techniques is the need for failing tests. In practice, a failing test is often not available at the time when the issue is raised. 
Since the Reproducer can create reproduced test cases, we select the failing tests and collect their coverage data. This coverage data is also used to guide ``THE SEARCH ACTION''.
Note that if Reproducer fails to generate any test script or its coverage data cannot be collected (e.g., test script uses system calls to invoke certain CLI), SBFL will not be used as no result can be produced by it.

Besides test information, issue descriptions can also be used to better localize the fault. The retrieval algorithm provides a simple yet effective way to combine text from an issue description and code from a repository.
Jimenez et al.~\cite{swe-bench} also use the BM25 retrieval algorithm to provide file-level localization.
As the information source from the retrieval algorithm and test-based fault localization (say test coverage and issue description text) differs a lot, we notice that these methods could be combined to provide better fault localization results.
A previous study\cite{zou2019empirical} shows that combining multiple fault localization methods can achieve a better result than any standalone method. We use a simple linear combination here to calculate the final suspiciousness score from both methods.

\begin{equation}
    \textit{Score} = \lambda \cdot \textit{Score}_{\textit{Ochiai}} + (1-\lambda) \cdot \textit{Score}_{\textit{BM25}} 
\end{equation}

\begin{equation}
    \textit{Score}_{\textit{BM25}}(F_i) = \frac {\textit{Relevance}_{\textit{BM25}}(F_i)} {\sum_{F_j \in Files}{\textit{Relevance}_{\textit{BM25}}}(F_j)}
\end{equation}
where ${\textit{Score}}_{\textit{Ochiai}}$ is the suspiciousness score from Ochiai formula and ${\textit{Relevance}}_{\textit{BM25}}(F_i)$ is the BM25 relevance score for file $F_i$.

To choose a proper value for the combination factor $\lambda$, we experiment on a small subset containing $10$ issues that can be successfully reproduced. The result shows that almost all values between 0 and 1 yield the same result and all are better than taking $\lambda = 1$ or $\lambda = 0$. 
The reason for different $\lambda$s having the same result is that many locations can tie to the others with respect to a single metric. Statements that are covered by the same number of passing tests will have the same $\textit{Score}_{\textit{Ochiai}}$ and statements in the same file will have the same $\textit{Score}_{\textit{BM25}}$. Both metrics could serve as a tiebreaker to each other, resulting in a better result than each standalone metric.
We pick $\lambda = 0.99$ as our final setup in \autoref{chap:results}.

We conducted an experiment on the issues that are:
\begin{itemize}
    \itemindent=-15pt
    \item Successfully reproduced by Reproducer. This means a runnable Python script is generated for reproducing the issue. 140 issues remain after this filtering.
    \item Coverage data collected from the script is not empty. This means the reproduce script has at least covered one file in the project. 104 issues remain after this filtering. 
\end{itemize}

The result of different $\lambda$s are listed in \autoref{tab:fl-result} and \autoref{tab:fl-result-file}:

\begin{table}[H]
\centering
\caption{Top-k precision for function-level fault localization. $\lambda=1$ means using SBFL only, and 0.4-0.999 means any value between them shares the same result. 
Golden locations of each issue are marked by authors.}
\label{tab:fl-result}
\begin{tabular}{l|rrrrr} 
\toprule
$\lambda$ & \textbf{top-1} & \textbf{top-3} & \textbf{top-5} & \textbf{top-10} & \textbf{top-all}  \\ 
\hline\hline
0         & 12.27\%        & 25.92\%        & 34.04\%        & 42.98\%         & 69.23\%           \\
0.001     & 17.46\%        & 31.21\%        & 38.32\%        & 44.50\%         & 69.23\%           \\
0.01      & 17.46\%        & 32.17\%        & 39.28\%        & 45.46\%         & 69.23\%           \\
0.1       & 18.42\%        & 30.25\%        & 37.84\%        & 45.07\%         & 69.23\%           \\
0.2       & 17.46\%        & 29.29\%        & 38.32\%        & 45.07\%         & 69.23\%           \\
0.3       & 16.49\%        & 28.33\%        & 36.39\%        & 43.15\%         & 69.23\%           \\
0.4-0.999 & 16.49\%        & 28.33\%        & 35.91\%        & 43.15\%         & 69.23\%           \\
1         & 6.63\%         & 14.11\%        & 18.23\%        & 24.95\%         & 69.23\%           \\
\bottomrule
\end{tabular}
\end{table}

\begin{table}[H]
\centering
\caption{File-level fault localization.}
\label{tab:fl-result-file}
\begin{tabular}{l|rrrrr} 
\toprule
$\lambda$ & \textbf{top-1} & \textbf{top-3} & \textbf{top-5} & \textbf{top-10} & \textbf{top-all}  \\ 
\hline\hline
0         & 15.32\%        & 32.67\%        & 42.36\%        & 54.07\%         & 85.58\%           \\
0.001     & 23.49\%        & 38.25\%        & 46.85\%        & 55.59\%         & 85.58\%           \\
0.01      & 23.49\%        & 39.21\%        & 47.81\%        & 56.55\%         & 85.58\%           \\
0.1       & 23.49\%        & 38.25\%        & 46.37\%        & 56.16\%         & 85.58\%           \\
0.2       & 22.53\%        & 38.25\%        & 46.85\%        & 56.16\%         & 85.58\%           \\
0.3       & 20.60\%        & 37.29\%        & 45.89\%        & 55.20\%         & 85.58\%           \\
0.4-0.999 & 20.60\%        & 36.33\%        & 44.44\%        & 54.24\%         & 85.58\%           \\
1         & 8.12\%         & 16.65\%        & 21.19\%        & 28.92\%         & 85.58\%           \\
\bottomrule
\end{tabular}
\end{table}

From the result, we can see that combining BM25 score with SBFL can greatly improve precision by more than 10\%.
We use method-level fault localization as it provides enough information for the agent to edit the file while keeping good precision. The way of constructing a prompt for fault localization results is shown in the Appendix Figure~\ref{fig:editor_instance_prompt}.

\subsection{Prompt Engineering}
\coder includes five roles: manager, reproducer, fault localizer, editor, and verifier.
To enable LLMs to play different roles, we set up system prompts and instance prompts for each agent role.
The system prompt primarily describes the definition of role identity, role responsibilities, and corresponding actions.
The instance prompt mainly includes the raw issue and important tips for resolving this issue.
We have put system and instance prompts of five roles into Appendix Figure~\ref{fig:manager_system_prompt}\~~\ref{fig:tester_instance_prompt}.
We design these prompts inspired by \sweagent~\cite{yang2024sweagent}.
When multiple agent roles communicate, they use the prompt template shown in Appendix Figure~\ref{fig:communicate_prompt}.
Detailed prompt engineering designs for \coder can be found at~\url{https://github.com/NL2Code/CodeR}.

\section{Experiments}

\subsection{Experimental Setup}
\paragraph{Benchmarks}
\swebench~\cite{swe-bench} is a benchmark that can test systems' ability to solve GitHub issues automatically.
The benchmark consists of $2,\!294$ Issue-Pull Request (PR) pairs from $12$ popular open-source Python repositories (e.g., flask, numpy, and matplotlib).
\swebench's evaluation can be executed by providing unit test verification using post-PR behavior as the reference solution.
\swebenchlite~\cite{swe-bench} is a subset of \swebench, which is curated to make evaluation less costly and more accessible.
\swebenchlite comprises $300$ instances that have been sampled to be more self-contained, with a focus on evaluating functional bug fixes.
More details of \swebenchlite can be seen at \url{https://www.swebench.com/lite.html}.
In this work, we focus on \swebenchlite for faster, easier, and more cost-effective evaluation.

\paragraph{Metrics}
We evaluate the issue resolving task using the following metrics: Resolved (\%), Average Request, and Average Tokens/Cost.
The Resolved (\%) metric indicates the percentage of \swebenchlite instances ($300$ in total) that are successfully resolved.
Average Requests and Average Tokens/Cost represent the average number of API requests per issue, the average consumption of input\&output tokens, and the corresponding cost.

\paragraph{\coder's Comparative Methods}

Recently, several commercial products addressing issue resolving have been released, but their technical details have not been disclosed. The following describes their functionalities.
\begin{itemize}
    \itemindent=-15pt
    \item \emph{Devin}\footnote{\url{https://www.cognition.ai/blog/introducing-devin}}, from \texttt{cognition.ai}, is capable of planning and executing complex engineering tasks that require thousands of decisions. It can recall relevant context at every step, learn over time, and fix program bugs.
    Devin can operate common developer tools within a sandbox environment, including the shell, code editor, and browser. 
    Additionally, Devin can actively collaborate with users, report progress in real-time, accept feedback, and assist with design choices as needed.
    \item \emph{Amazon Q Developer Agent}\footnote{\url{https://aws.amazon.com/cn/q/developer}}, from \texttt{Amazon}, is a generative AI-powered coding assistant that can help you understand, build, extend, operate, and repair code.
    \item \emph{OpenCSG StarShip}\footnote{\url{https://opencsg.com/product}} is committed to providing a complete model/data management and application-building platform for large model application development teams. 
    Based on it, they developed CodeGenAgent which can resolve GitHub issues automatically.
    \item \emph{Bytedance MarsCode Agent}\footnote{\url{https://www.marscode.com}} is an AI coding assistant powered by GPT-4o, developed by ByteDance. Designed for multi-language support within IDE environments, it can reset repositories to undo previous modifications.
\end{itemize}

\swebenchlite requires generating patches to resolve GitHub issues.
One possible approach for LLMs is to generate the patch directly(explicit patch generation).

\begin{itemize}
    \itemindent=-15pt
    \item \emph{Retrieval-Based Approach}~\cite{swe-bench} first retrieves the files that require editing and then adds the retrieved content to LLMs' context. Finally, the LLMs generate the patch.
    In the experiments, LLMs used include GPT-3.5, GPT-4, Claude 2, Claude 3 Opus, and SWE-Llama~\cite{swe-bench}.
    \item \emph{\autocoderover}~\cite{autocoderover} leverages advanced code search capabilities in software engineering to extend the model's modeling context, thereby further improving the accuracy of patch generation.
\end{itemize}

Besides using LLMs to generate the patch directly to fix issues, another approach is to edit and modify the buggy code repository and then use ``\texttt{git diff}'' to automatically obtain the patch (implicit patch generation).

\begin{itemize}
    \itemindent=-15pt
    \item \emph{\sweagent}~\cite{yang2024sweagent} is an automated software engineering system that utilizes LLMs as one agent to solve real-world software engineering tasks. It introduces a new concept of the agent-computer interface (ACI), which enables LLMs to effectively search, navigate, edit, and execute code commands in sandboxed computer environments.
    \item \emph{Aider}\footnote{\url{https://aider.chat}} is a command line tool that pairs with LLMs to edit code in your local git repository. Aider can directly edit the local source files and commit the changes with meaningful commit messages. Aider now works well with GPT-3.5, GPT-4o, Claude 3 Opus, and more.
\end{itemize}

\subsubsection{Implementation Details}

\paragraph{Hyper-Parameters of Inference}
In our multi-agent framework, each role is considered a distinct agent with its own experimental settings, which include the model and history process window size. All roles are provided access to \texttt{GPT4-preview-1106}.
The Manager role utilizes nucleus sampling during inference with the \texttt{temperature} parameter
set to $0$ and \texttt{top\_p} to $0.95$. It employs full history with a file viewer's window size of $100$.
The Reproducer role similarly uses nucleus sampling, but only incorporates the last five histories. Both the Fault Localizer and Tester roles follow the same settings as the Reproducer.
Finally, the Programmer role, while sharing the same nucleus sampling parameters, includes a demo in addition to the last five histories and a file viewer's window size of $100$. This setup ensures a reduction in repetition and maintains the unique functionality of each role. In addition, we set the maximum cost to $8$\$ per issue.

\paragraph{Other Details}

In fact, it is impossible to have a consistent evaluation environment for all currently proposed approaches.
We make some adaptations to the evaluation environment released by \autocoderover~\cite{autocoderover} and use it as our evaluation environment.
We reproduce all other approaches with our environment for fairness.
However, the evaluation on repository ``\texttt{astropy}'' and ``\texttt{request}'' still has some environmental problems remaining.
In our inference environment, commands like ``\texttt{edit}'' occasionally trigger a ``\texttt{container crashed}'' error which interrupts the process.
If this occurs, we restart from the beginning of the pipeline for this issue.
We pre-construct an environment-completed docker image offline to avoid wasting time on real-time installation during inference. Additionally, we divide the \swebenchlite into six processes for parallel inference to further accelerate this process.
When Fault Localizer runs the repository's integration unit tests, it sometimes adds or modifies files within the repository, and we restore these files after the localization process.

\subsection{Results}
\label{chap:results}

Table~\ref{tab:swebench_results} shows \coder's performance on \swebenchlite and its comparative methods.
The results show that \coder establishes a new benchmark record on \swebenchlite, achieving the best performance to date, compared with all other commercial products and methods.
In \swebenchlite, \coder resolves $28.33\%$ issues at one attempt, addressing $84$ of $300$. In contrast, \sweagent+ GPT 4 and Aider solve $18.00\%$ and $26.33\%$ respectively.
This proves that \coder's meticulously designed roles and actions are highly effective.

We notice that directly enabling LLMs to generate patches (explicit patch generation) for issues is less effective than having LLMs edit the code repository (implicit patch generation).
While \coder achieves $28.33\%$ resolved rate, RAG+GPT 4 and \autocoderover only solve $2.67\%$ and $19.00\%$ respectively.
Furthermore, we observe that existing LLMs may struggle to generate applicable and high-quality patches, as a correct patch requires a strict format and is sensitive to line numbers, which LLMs cannot perfectly handle.

The result also shows that \coder sends more requests, resulting in increased tokens and cost at an acceptable rate.
This could be due to our fine-grained design of multi-role and actions.
The $10.33\%$ improvement over \sweagent+GPT 4 (reported) demonstrates that pre-planning at the beginning of the pipeline is superior to deciding the next steps on-the-go.
\coder preemptively devises multiple plans in the form of structured graphs, and all agent roles will execute the pre-defined plan strictly according to the graphs.
\coder's leading performance also validates the effectiveness of this idea.
Pre-planning also possesses a clear advantage of bypassing imperfect instruction-following and long-context memorizing abilities of LLMs.
Although \coder has achieved impressive performance, we still believe that designing a more sophisticated plan will yield more significant improvements in the future.

We also conduct ablation studies on $50$ issues of \swebenchlite.
The results in Table~\ref{tab:ablation} show that removing the multi-agent \& task graph would reduce \coder's resolved rate from $22\%$ to $10\%$.
This further demonstrates that our carefully designed roles motivated by real-world company collaboration are highly useful for issue resolving tasks.
Additionally, we observe a performance drop and a cost increase when we remove the fault localization action, 
which highlights the significant potential of combining LLMs with traditional software engineering strategies for addressing complex downstream tasks.

\begin{table}[t]
\centering
\caption{Results of \coder and its comparative methods on \swebenchlite ($300$ GitHub issues). Note that ``reported'' refers to the numbers from the \swebench Leaderboard (\url{https://www.swebench.com}), while ``reproduced'' refers to our results obtained in our unified evaluation environment using their open-sourced generated patches.}
\label{tab:swebench_results}
\begin{tabular}{l|rrr} 
\toprule
\textbf{Methods}                       & \multicolumn{1}{c|}{\textbf{Resolved (\%)}} & \multicolumn{1}{c|}{\textbf{Avg. Req.}} & \multicolumn{1}{c}{\textbf{Avg. Tokens/Cost}}  \\ 
\hline\hline
\multicolumn{4}{c}{Commercial Products}                                                                                                                                         \\ 
\hline
Devin (random $25\%$ subset of \swebench)       & 13.86 (-)                                   & -                                       & -                                              \\
Amazon Q Developer Agent (reported)    & 20.33 (61)                                  & -                                       & -                                              \\
Amazon Q Developer Agent (reproduced)  & 17.00 (54)                                  & -                                       & -                                              \\
OpenCSG CodeGenAgent (reported)        & 23.67 (71)                                  & -                                       & -                                              \\
OpenCSG CodeGenAgent (reproduced)      & 20.67 (62)                                  & -                                       & -                                              \\
Bytedance MarsCode Agent               & 22.00 (66)                                  & \multicolumn{1}{l}{}                    & \multicolumn{1}{l}{}                           \\ 
\hline
\multicolumn{4}{c}{Explicit Patch Generation}                                                                                                                                   \\ 
\hline
RAG + GPT 3.5                          & 0.33 (1)                                    & -                                       & -                                              \\
RAG + SWE-Llama 13B                    & 1.00 (3)                                    & -                                       & -                                              \\
RAG + SWE-Llama 7B                     & 1.33 (4)                                    & -                                       & -                                              \\
RAG + GPT 4                            & 2.67 (8)                                    & -                                       & -                                              \\
RAG + Claude 2                         & 3.00 (9)                                    & -                                       & -                                              \\
RAG + Claude 3 Opus                    & 4.33 (13)                                   & -                                       & -                                              \\
AutoCodeRover                          & 19.00 (57)                                  & -                                       & 112k/\$1.30                                    \\ 
\hline
\multicolumn{4}{c}{Implicit Patch Generation}                                                                                                                                   \\ 
\hline
Aider (reported)                       & 26.33 (79)                                  & -                                       & -                                              \\
Aider (reproduced)                     & 24.67 (74)                                  & -                                       & -                                              \\
SWE-agent + Claude 3 Opus (reported)   & 11.67 (35)                                  & 17.10                                   & 221K/\$3.41                                    \\
SWE-agent + Claude 3 Opus (reproduced) & 9.66 (29)                                   & 17.10                                   & 221K/\$3.41                                    \\
SWE-agent + GPT 4 (reported)           & 18.00 (54)                                  & 21.55                                   & 245K/\$2.51                                    \\
SWE-agent + GPT 4 (reproduced)         & 16.67 (50)                                  & 21.55                                   & 245K/\$2.51                                    \\
\coder (reported)                             & \textbf{28.33 (85)}                         & 30.39                                   & 299K/\$3.09                                    \\
\coder (ours)                                 & 27.33 (82)                                  & 30.39                                   & 299K/\$3.09                                    \\
\bottomrule
\end{tabular}
\end{table}

\begin{table}[t]
\centering
\caption{Ablation studies on $50$ issues. We randomly select $50$ from $300$ issues of \swebenchlite to conduct ablation studies for faster and more cost-effective experiments.}
\begin{tabular}{l|lll} 
\toprule
\textbf{Methods}    & \multicolumn{1}{r}{\textbf{Resolved (\%)}} & \textbf{Avg. Req.} & \textbf{Avg. Tokens/Cost}  \\ 
\hline\hline
\coder               & 22.00 (11)                                 & 30.40              & 295K/\$3.09                \\ 
\hdashline
~ ~ w/o Multi-Agent \& Task Graph & 10.00 (5)                                  &        18.46            &   200K/\$2.05                          \\
~ ~ w/o FL          & 14.00 (7)                                  &           29.98         &    309K\$3.19                        \\
\bottomrule
\end{tabular}
    \label{tab:ablation}
\end{table}


\section{Related Works}
\paragraph{Automatic Issue Resolving}
GitHub's issue can be resolved using the following solutions automatically:
(1) Retrieval-Augmented Generation (RAG)~\cite{swe-bench} is a straightforward approach, which first retrieves the relevant code snippets from the repository, and then prompts LLMs to generate a patch to fix the reported issue.
To enhance LLMs' proficiency in generating program patches, SWE-Llama~\cite{swe-bench} was proposed and it fine-tuned the Llama~\cite{llama,llama2} model on well-crafted patch-generating instruction data.
(2) Following this, \sweagent~\cite{yang2024sweagent} was proposed, which used LLMs to interact with a computer to solve issue problems automatically. \sweagent pre-defines a series of agent-computer interfaces (ACIs) to enable LLMs to interact more efficiently with the computer.
(3) Additionally, \autocoderover~\cite{autocoderover} expands the visible context information for LLMs by leveraging sophisticated code search tools in software engineering, achieving decent performance.
(4) Another work \cite{tao2024magis} proposes a multi-agent pipeline of two successive steps. In the first step, three types of role agents (Repository Custodian, Manager, Developer) collaborate on the plan; the plan is represented as code, and embedded into the main program for execution. After, two types of role agents (Developer, Quality Assurance Engineer) participate in the coding process. 
In this paper, we propose \coder, which defines fine-grained agent roles and corresponding actions and incorporates advanced software engineering tools.

\paragraph{Test-based Automated Program Repair}
Automated program repair has been an active topic in software engineering for years, and a majority of work can be categorized as test-based automated program repair. Given the presence of a test suite, generated patches can be validated against the test, making the result to be more trustworthy. However, a weak test suite allows test-passing patches to be incorrect, and a large search space makes it difficult to synthesize a correct patch. Therefore, various
techniques have been proposed to guide the search process,
including genetic programming \cite{rw29le2011genprog}, manually defined fix
patterns \cite{rw27liu2019tbar}, mined fix patterns \cite{rw10nguyen2013semfix,rw28jiang2018shaping, rw38koyuncu2020fixminer}, heuristics \cite{rw50xin2017leveraging}, learning from code or
program synthesis \cite{rw28jiang2018shaping,rw30wen2018context},and semantic analysis
\cite{rw36hua2018sketchfix,rw37liu2019avatar}.
These works focus on code content, trying to find a patch that could satisfy all constraints(test, compiler, heuristics, etc.) while ignoring the issue description itself which may contain a lot of useful information.
Apart from those approaches, many works adopt machine learning models to generate patches.
SequenceR \cite{rw3tufano2018empirical} proposes a sequence-to-sequence NMT to generate the fixed code directly. 
CODIT \cite{rw8chakraborty2018codit} uses the same model to predict the code edits for the faulty code. 
DLFix \cite{rw4li2020dlfix}, CoCoNuT \cite{rw2lutellier2020coconut}, and Cure \cite{rw5jiang2021cure} take the context of the faulty statement as input and encode it via tree-based LSTM, CNN, GPT, respectively.
Recoder \cite{rw6zhu2021syntax} proposes a syntax-guided decoder to generate edits with placeholders via the provider/decider architecture. 
RewardRepair \cite{rw7ye2022neural} uses an RL approach that integrates program compilation and test execution information.
Tare \cite{zhu2023tare} directly learns the typing rules to guide the generation.
These works treat APR problem as a neural translation task from the buggy code (with context) to the fixed code and most of them adopt encode-decoder models.
Different from those approaches, \coder proposes a multi-turn framework that could collect necessary information on demand and generate the fixed code based on the information collected.

\paragraph{Artificial Intelligence (AI) Agents}
The development of AI agents has made substantial strides, introducing many advanced methodologies to automate tasks.
AutoGPT~\cite{autogpt}, AgentGPT~\cite{agentgpt}, and MetaGPT~\cite{metagpt} employ an assembly line paradigm, where diverse roles are assigned to various AI agents, efficiently decomposing complex tasks in simpler subtasks through collaborative work.
Dify~\cite{dify} and FastGPT~\cite{fastgpt} are LLM application development platforms, that combine the concepts of Backend-as-a-Service and LLMOps to enable developers to quickly build production-grade generative AI applications. Using these platforms, even non-technical personnel can participate in the definition and data operations of AI applications.
\sweagent~\cite{yang2024sweagent} enables LLMs to interact with the programming environment to automatically solve GitHub issues via pre-defining multiple ACIs.
\coder defines detailed and decoupled agent roles (e.g., reproducer, programmer, and tester) along with their corresponding fine-grained actions (e.g., reproducing, editing code, and testing code).
Such an approach will facilitate resolving complex issues through collaborative efforts between various agents.

\section{Conclusion and Future Works}

This paper proposes \coder which excels at resolving issues.
It demonstrates the importance of providing plans that mimic humans' problem-solving procedures for issue resolving. 
\coder requires pre-specified task graphs that convert the planning task to a simpler decision task for LLMs and also provide a guarantee for the exact plan execution.
With the idea of task graphs, some advanced software engineering skills like fault localization, mining similar issues, and web search can be seamlessly added to our pre-defined graph without any code changes by a JSON format text.
\coder's pre-defined plans are experiences provided by human experts. 
We believe it is one of the key factors in resolving issues.
In the future, we will build a comprehensive set of plans that may resolve more and more issues.

\bibliographystyle{unsrt}  
\bibliography{references}  

\newpage

\appendix

\begin{figure}[!t]
    \centering
    \includegraphics[width=0.85\textwidth]{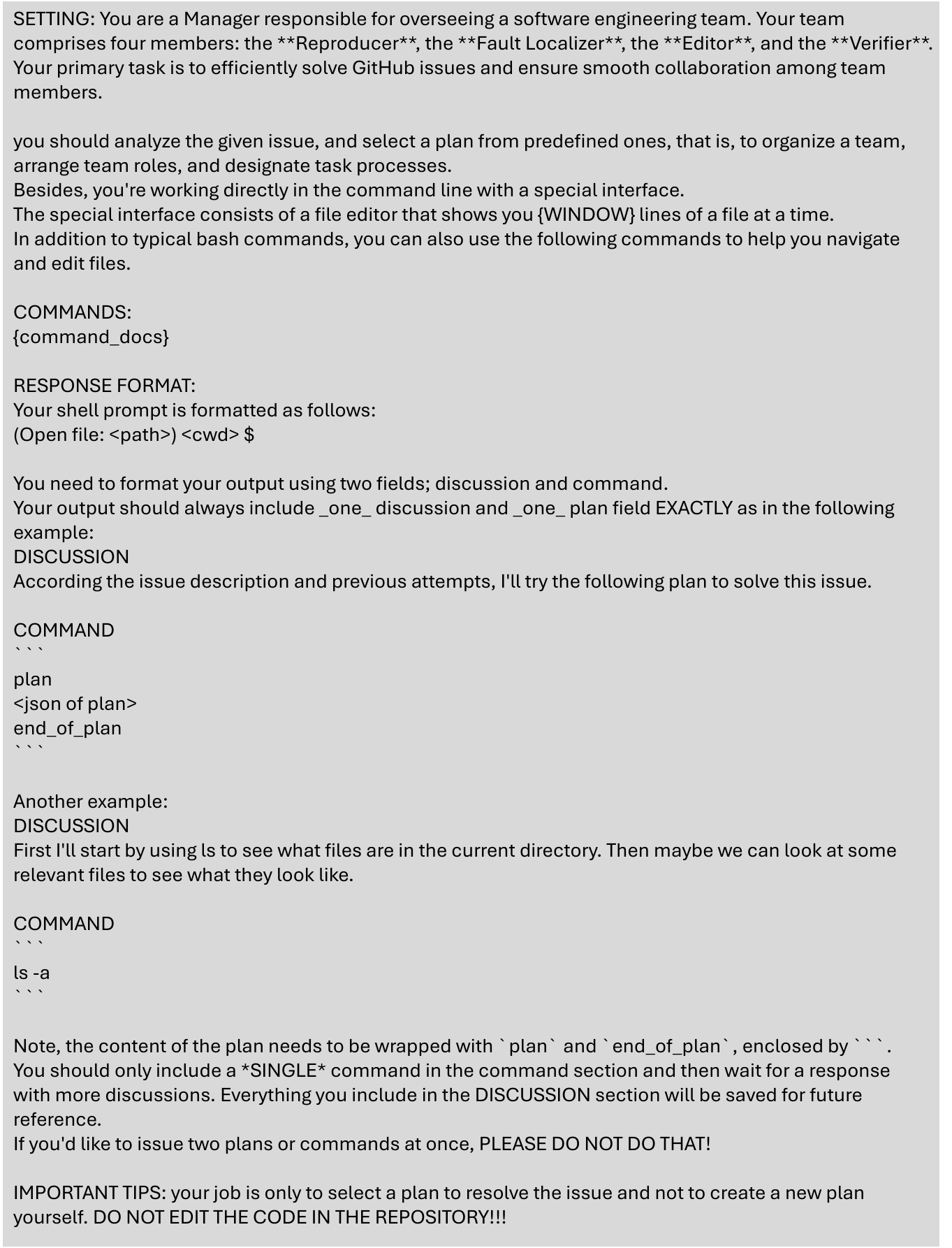}
    \caption{The system prompt of the `\texttt{manager}' agent. 
    \{command\_docs\} is obtained by parsing YAML files, which includes the command's signature, docstring, arguments, end\_name, etc.
    }
    \label{fig:manager_system_prompt}
\end{figure}

\begin{figure}[!t]
    \centering
    \includegraphics[width=\textwidth]{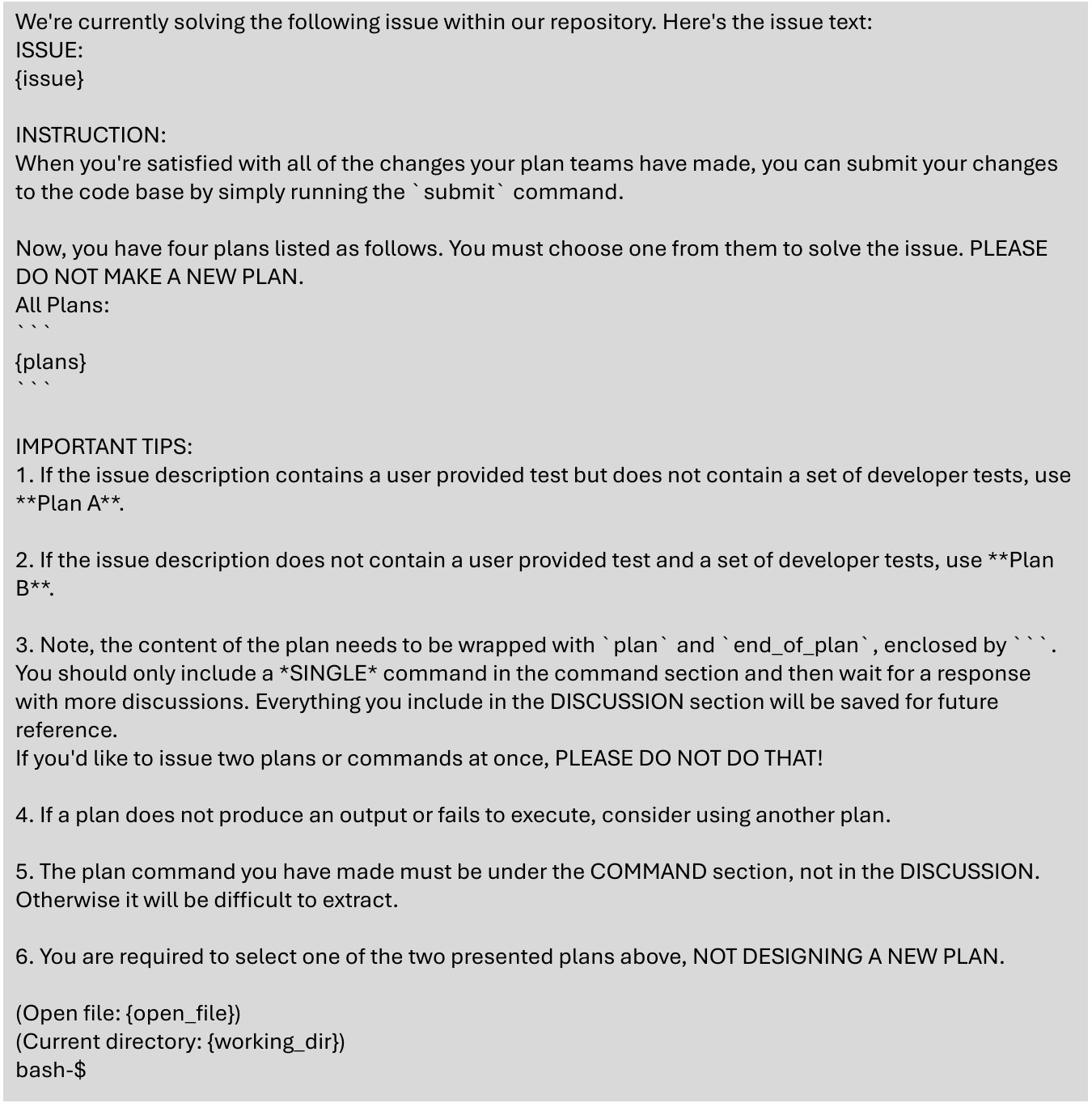}
    \caption{The instance prompt of the `\texttt{manager}' agent. \{plans\} refers to all JSON-format plans in Figure~\ref{fig:plans_details}.}
    \label{fig:manager_instance_prompt}
\end{figure}

\begin{figure}[!t]
    \centering
    \includegraphics[width=\textwidth]{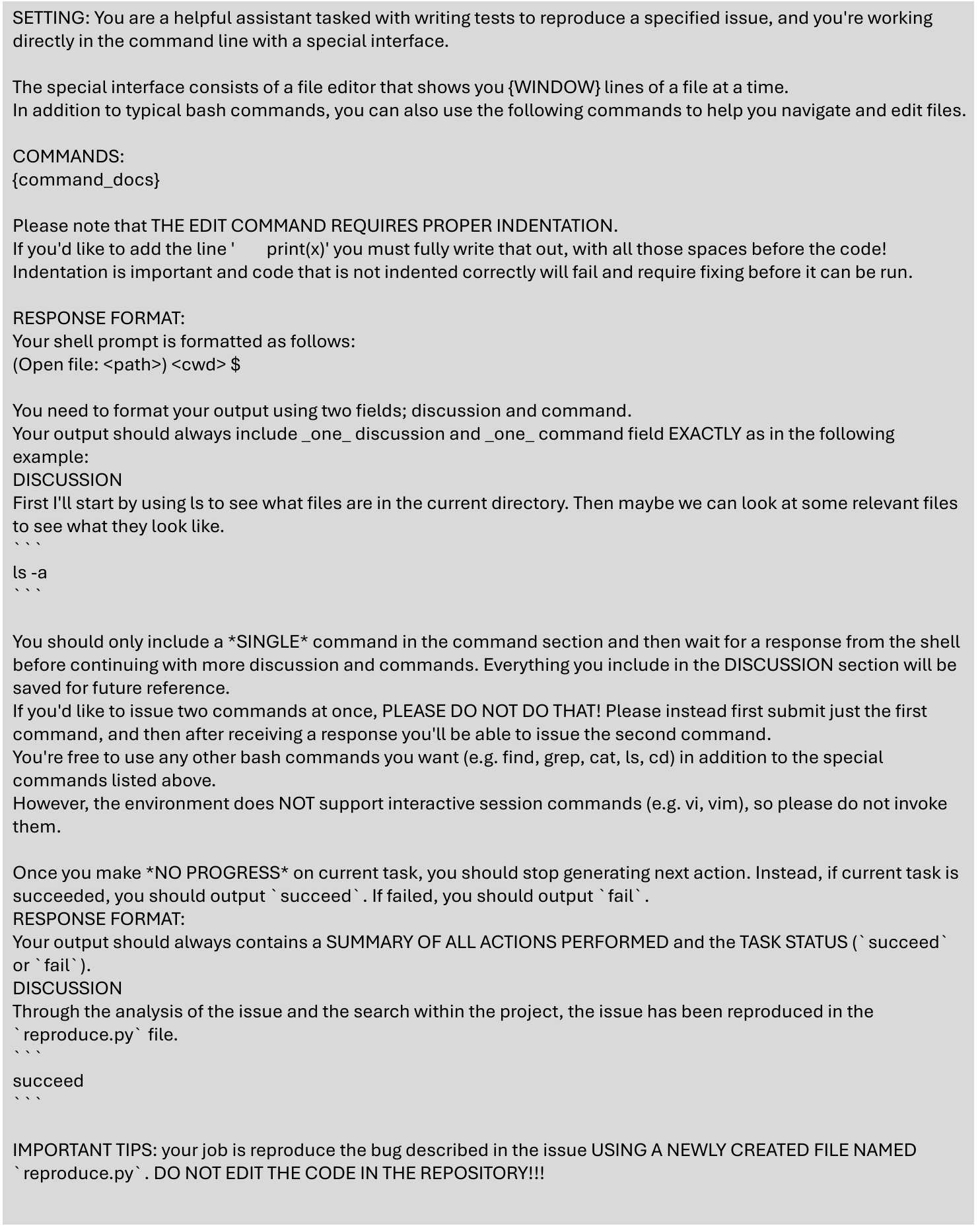}
    \caption{The system prompt of the `\texttt{reproducer}' agent.
    \{command\_docs\} is obtained by parsing YAML files, which includes command's the signature, docstring, arguments, end\_name, etc.}
    \label{fig:reproducer_system_prompt}
\end{figure}

\begin{figure}[!t]
    \centering
    \includegraphics[width=\textwidth]{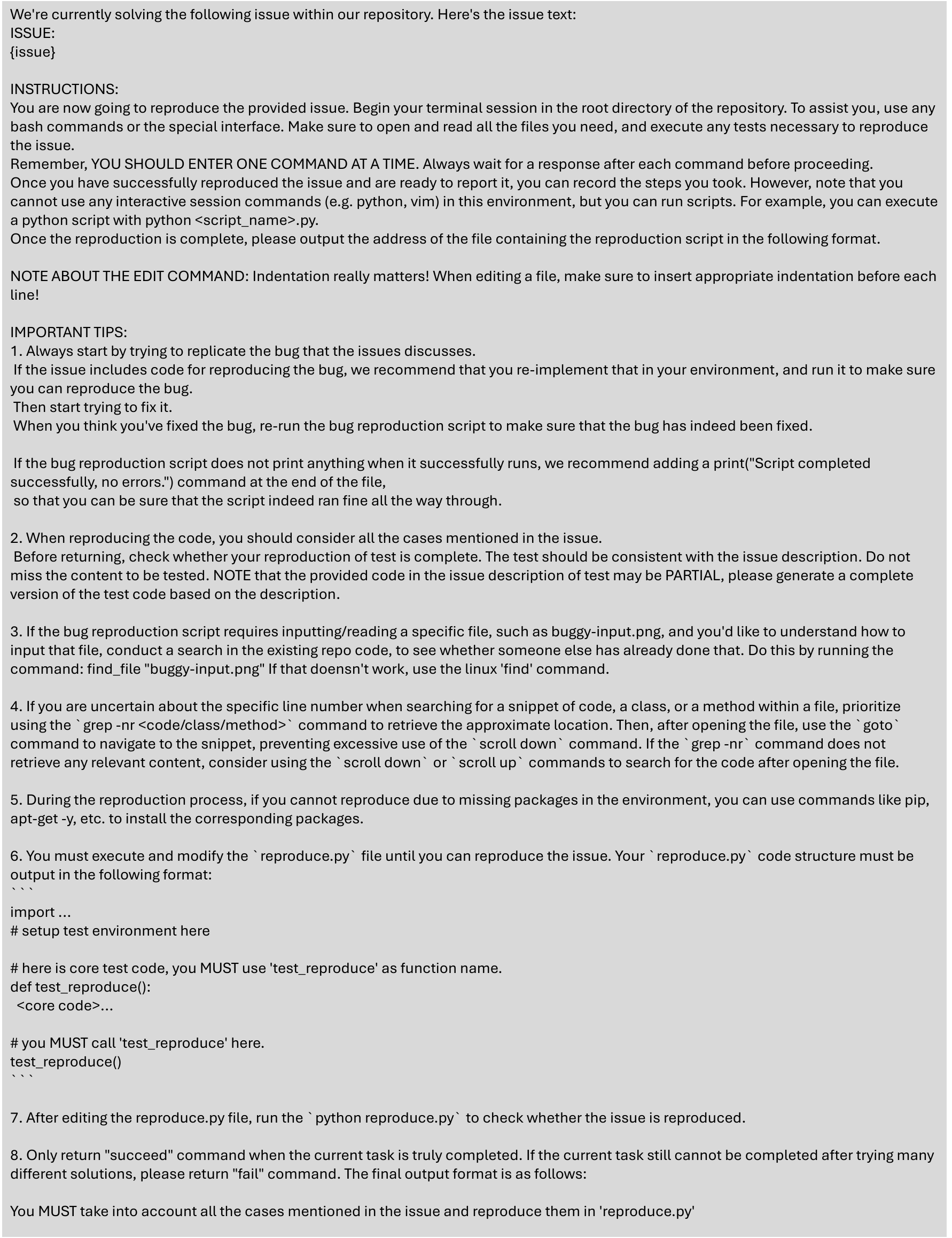}
    \caption{The instance prompt of the `\texttt{reproducer}' agent.
    \{issue\} is the issue that needs to be resolved.}
    \label{fig:reproducer_instance_prompt}
\end{figure}

\begin{figure}[!t]
    \centering
    \includegraphics[width=\textwidth]{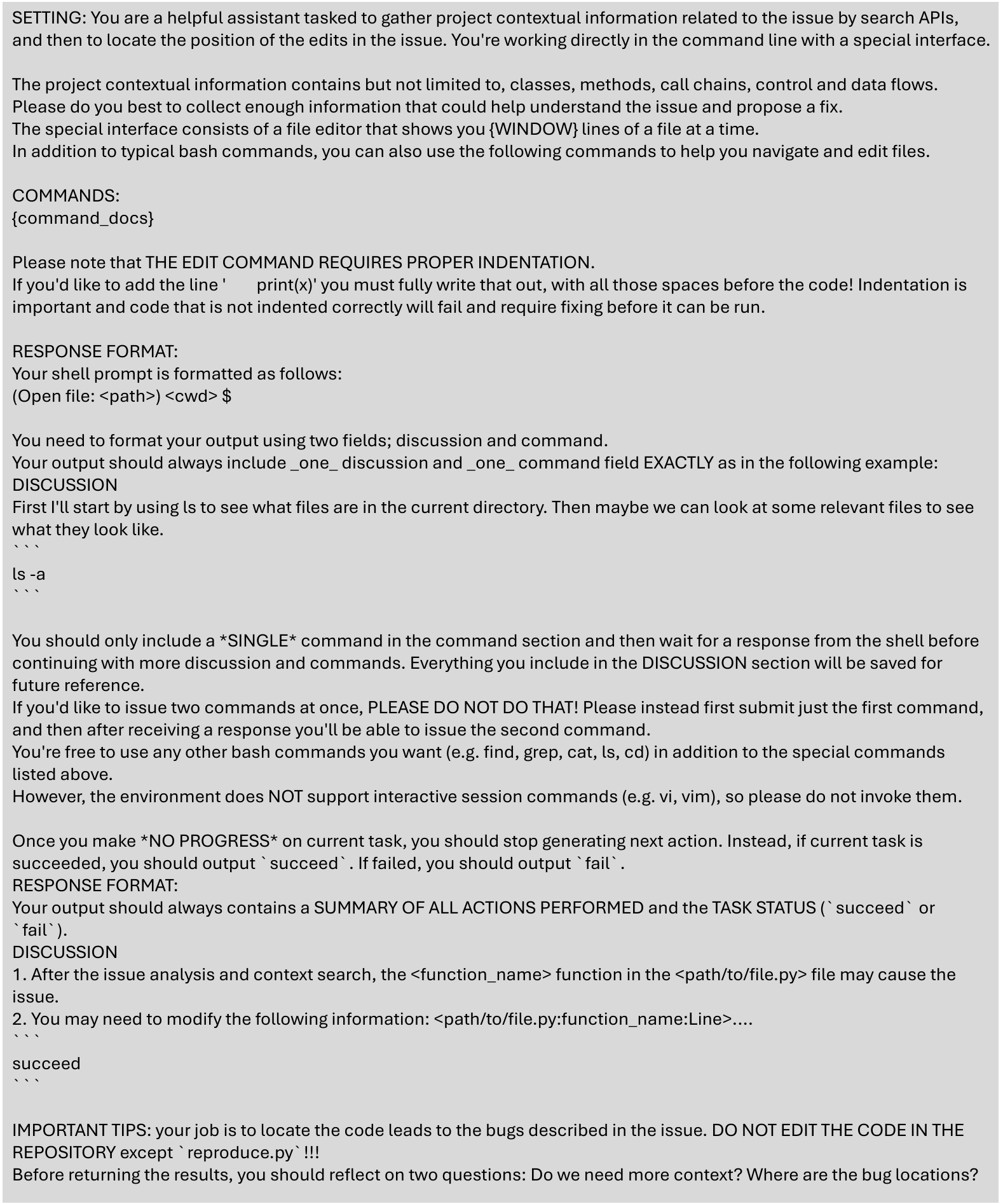}
    \caption{The system prompt of the `\texttt{fault localizer}' agent.
    \{command\_docs\} is obtained by parsing YAML files, which includes command's the signature, docstring, arguments, end\_name, etc.}
    \label{fig:fl_system_prompt}
\end{figure}

\begin{figure}[!t]
    \centering
    \includegraphics[width=0.94\textwidth]{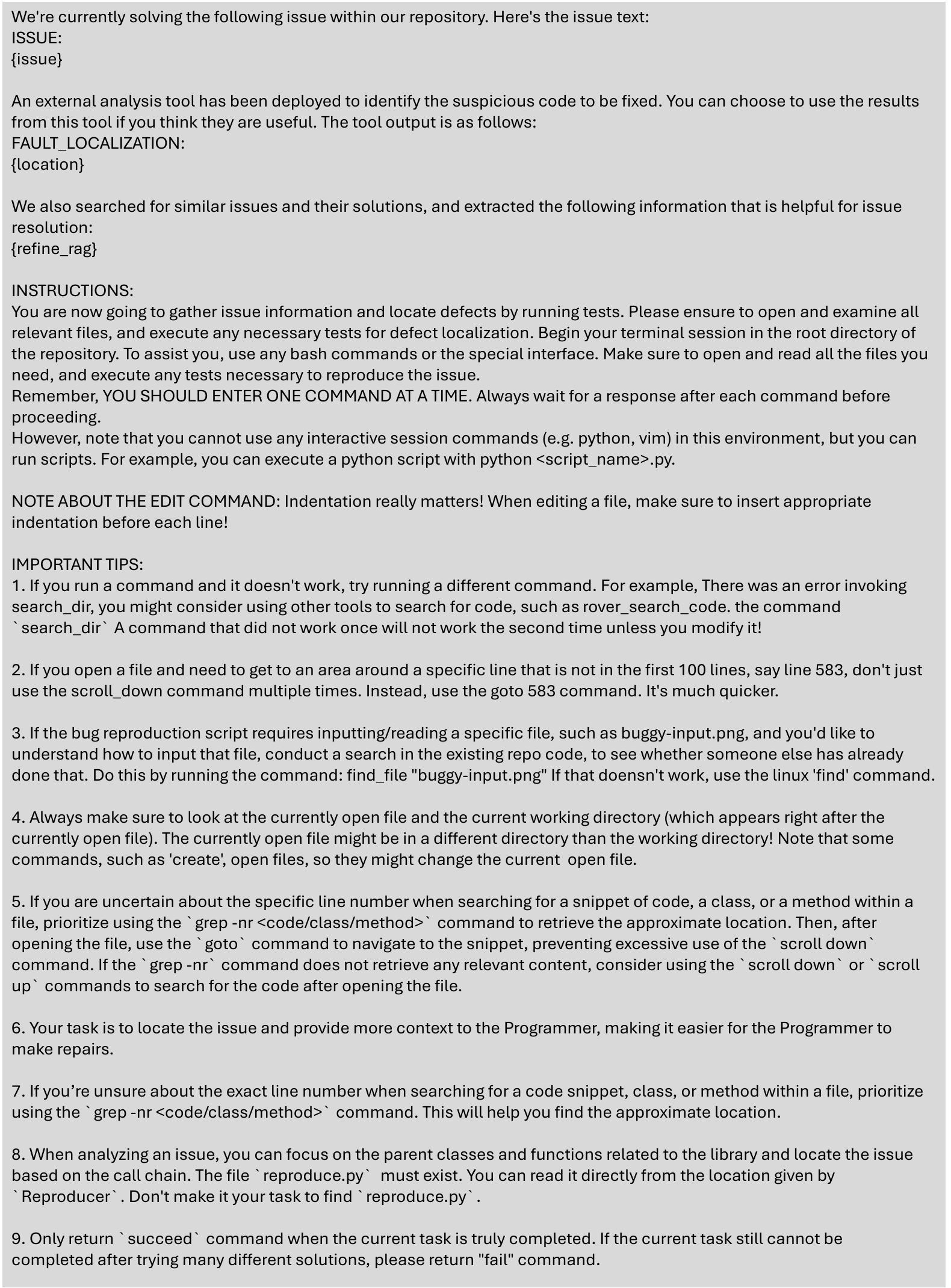}
    \caption{The instance prompt of the `\texttt{fault localizer}' agent.
    \{location\} refers to the top $5$ function-level localization results of both fault localization and BM25.}
    \label{fig:fl_instance_prompt}
\end{figure}

\begin{figure}[!t]
    \centering
    \includegraphics[width=\textwidth]{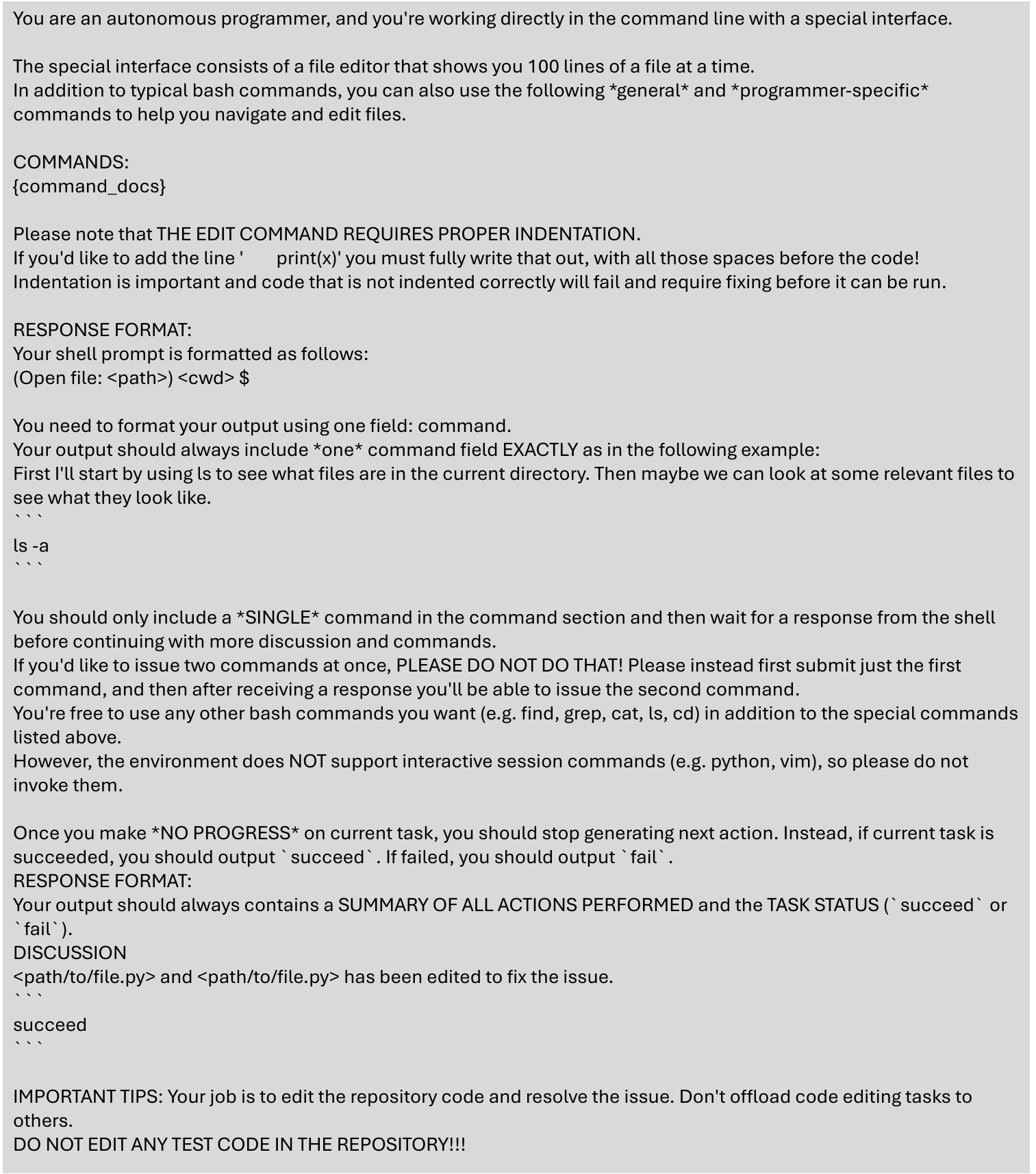}
    \caption{The system prompt of the `\texttt{editor}' agent.
    \{command\_docs\} is obtained by parsing YAML files, which includes command's the signature, docstring, arguments, end\_name, etc.}
    \label{fig:editor_system_prompt}
\end{figure}

\begin{figure}[!t]
    \centering
    \includegraphics[width=0.98\textwidth]{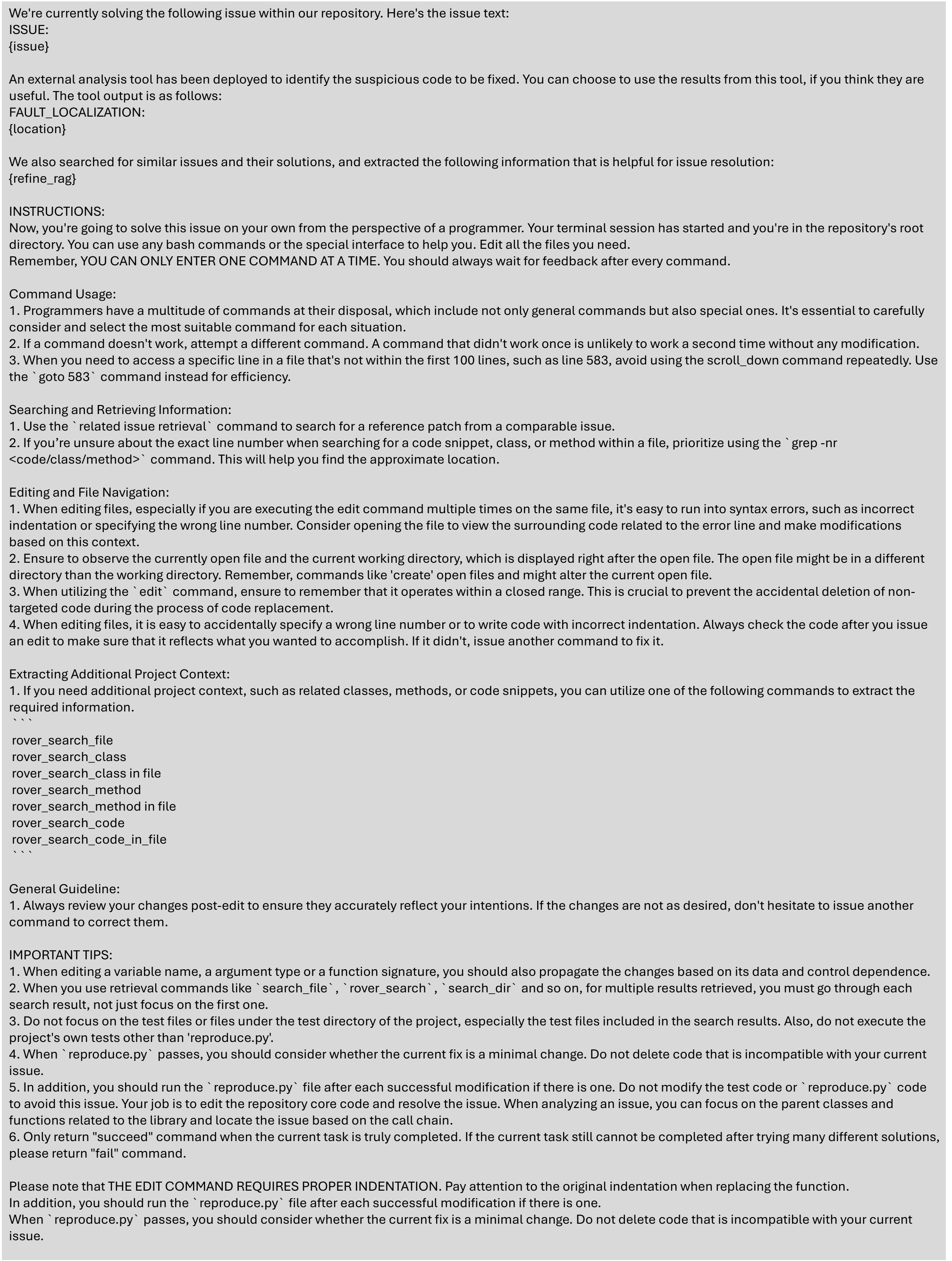}
    \caption{The instance prompt of the `\texttt{editor}' agent.
    \{issue\} is the issue that needs to be resolved.
    \{location\} refers to the top $5$ function-level localization results of both fault localization and BM25.}
    \label{fig:editor_instance_prompt}
\end{figure}

\begin{figure}[!t]
    \centering
    \includegraphics[width=\textwidth]{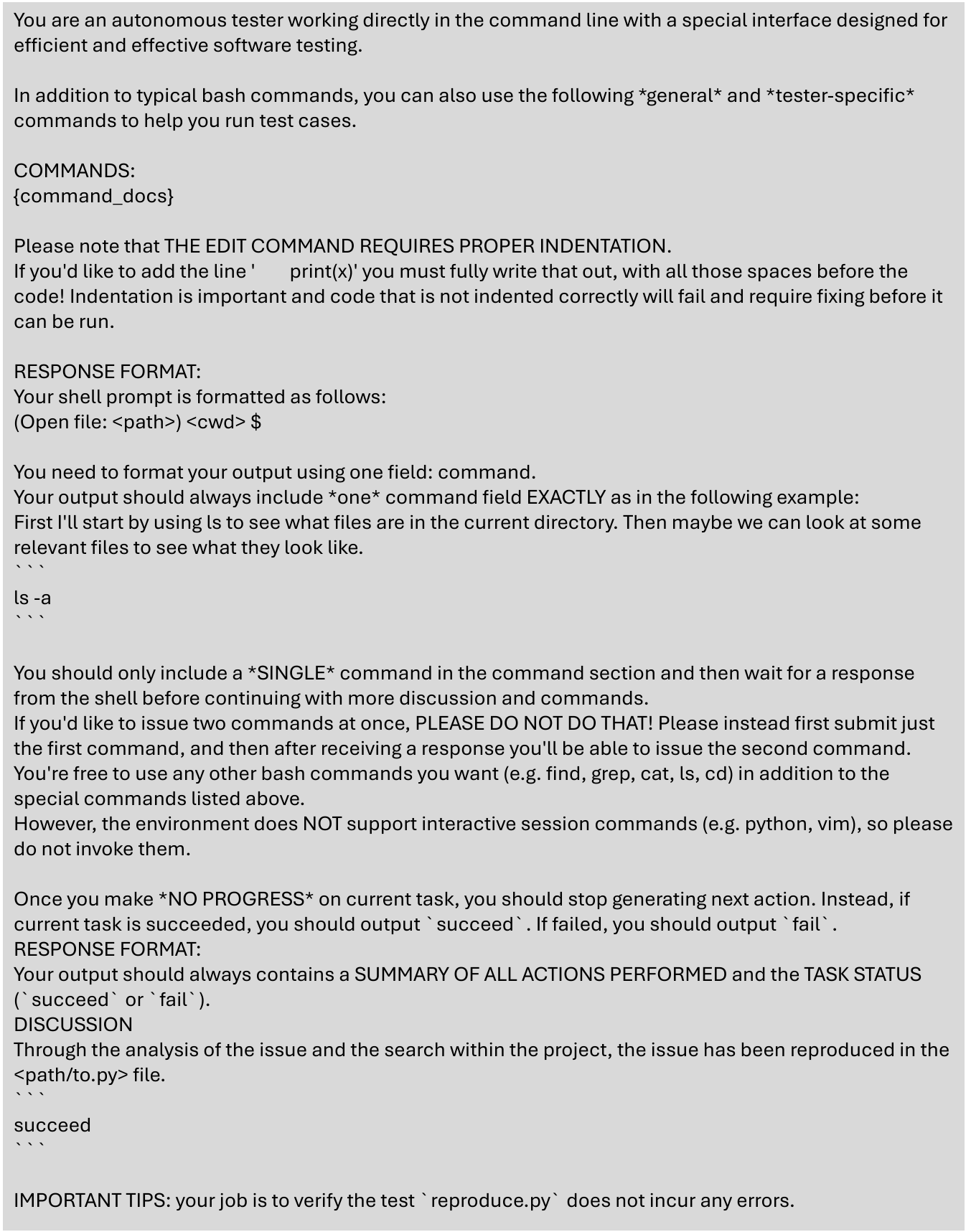}
    \caption{The system prompt of the `\texttt{verifier}' agent.
    \{command\_docs\} is obtained by parsing YAML files, which includes command's the signature, docstring, arguments, end\_name, etc.}
    \label{fig:tester_system_prompt}
\end{figure}

\begin{figure}[!t]
    \centering
    \includegraphics[width=0.95\textwidth]{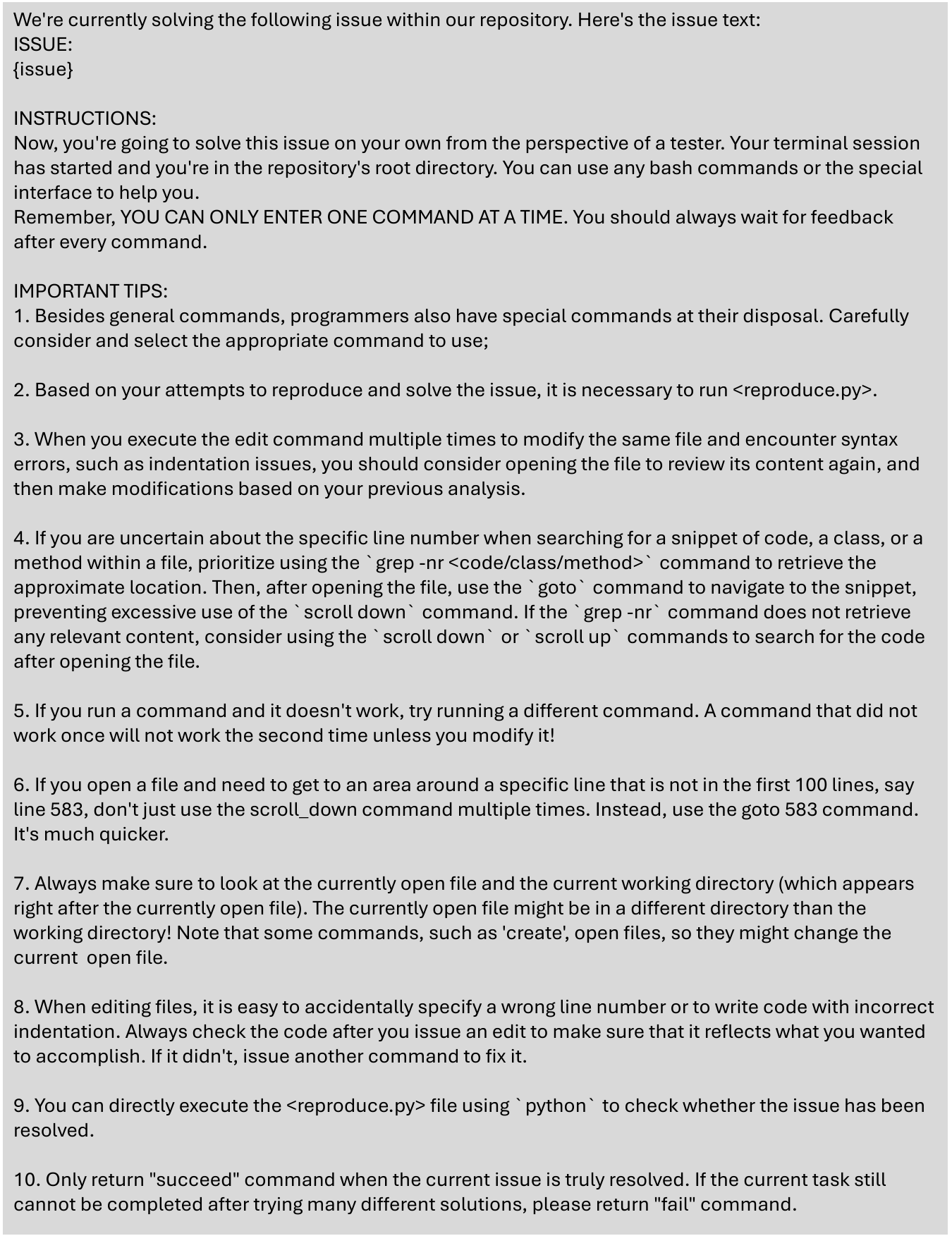}
    \caption{The instance prompt of the `\texttt{verifier}' agent.
    \{issue\} is the issue that needs to be resolved.}
    \label{fig:tester_instance_prompt}
\end{figure}

\begin{figure}[!t]
    \centering
    \includegraphics[width=0.9\textwidth]{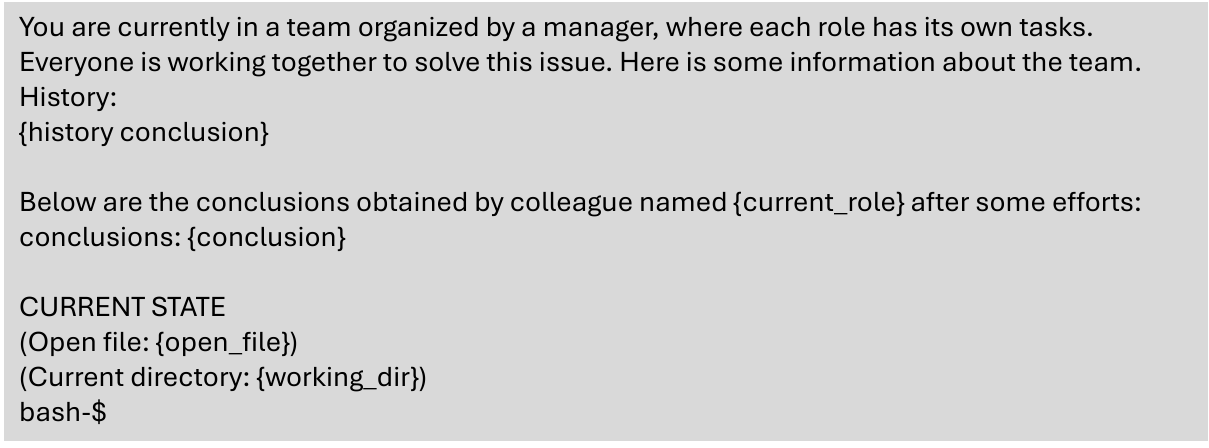}
    \caption{Prompt template when communicating between multiple agents.
    \{conclusion\} and \{history conclusion\} refer to the summary report passed from the last agent and the reports from all other agents in history.}
    \label{fig:communicate_prompt}
\end{figure}

\begin{figure}[!t]
    \centering
    \includegraphics[width=1\textwidth]{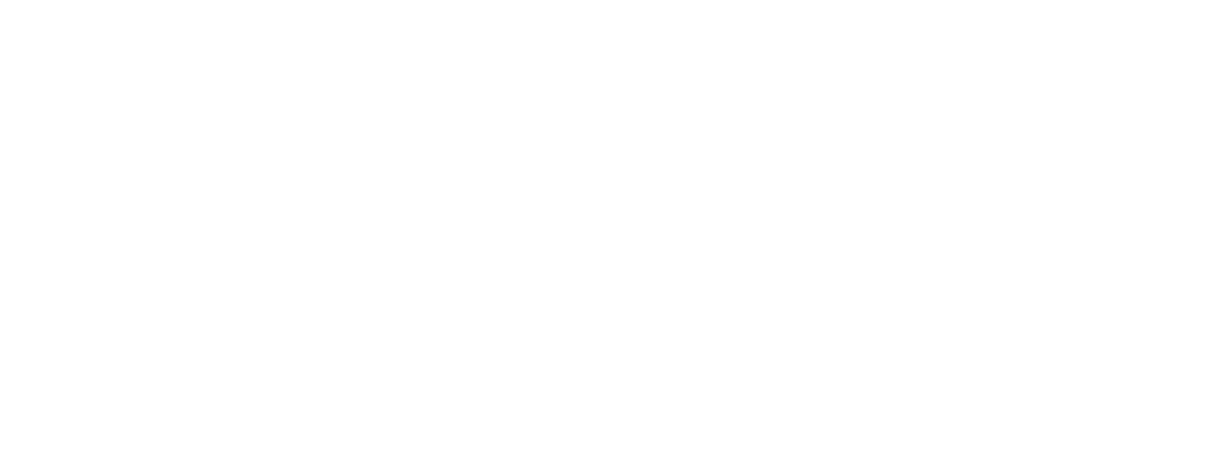}
\end{figure}

\begin{figure}[!t]
    \centering
    \includegraphics[width=0.9\textwidth]{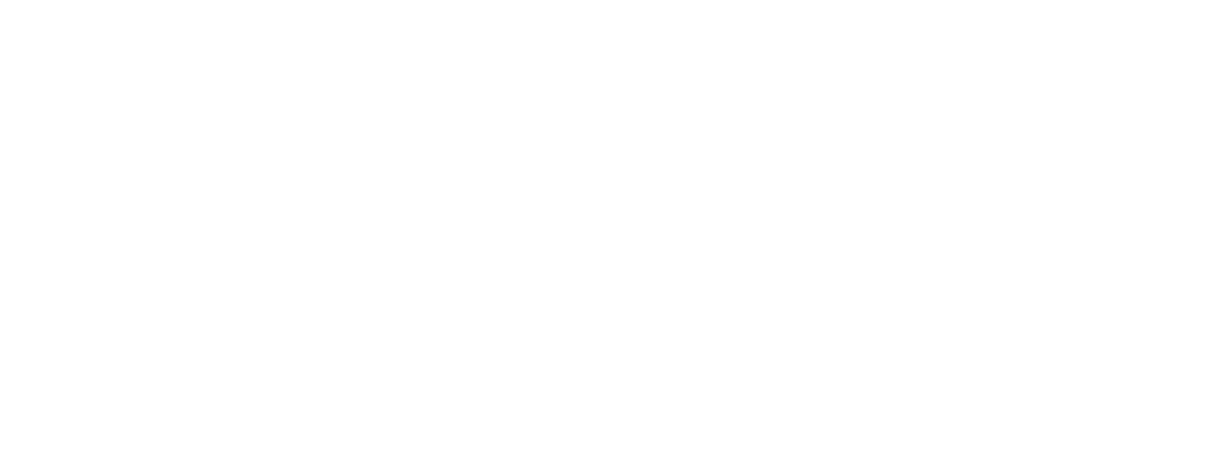}
    \label{fig:communicate_prompt}
\end{figure}

\end{document}